\documentclass[10pt,conference]{IEEEtran} 
\usepackage{cite}
\usepackage{amsmath,amssymb,amsfonts}
\usepackage{algorithmic}
\usepackage{graphicx}
\usepackage{textcomp}
\usepackage{xcolor}
\usepackage{multirow}  
\usepackage{multicol} 
\usepackage{booktabs}
\usepackage{subfigure}

\IEEEoverridecommandlockouts                              

\title{\LARGE \bf
MamMA: A Mamba-Based Pedestrian Trajectory Prediction Algorithm Considering Occupancy Map and Pedestrian Awareness States}

\author{Juncen Long, Xiaofeng Jin, Gianluca Bardaro, Simone Mentasti, Matteo Matteucci
\thanks{The authors are with the Department of Electronics, Information and Bioengineering, Politecnico di Milano, Milan, Italy. Email: \{juncen.long; xiaofeng.jin; gianluca.bardaro; simone.mentasti; matteo.matteucci\}@polimi.it.}%
}

\begin{document}

\maketitle
\thispagestyle{empty}
\pagestyle{empty}

\begin{abstract}

Many pedestrian trajectory prediction algorithms have been proposed to improve the safety of navigation for mobile robots working in human-robot coexistence environments. Some pedestrian trajectory prediction algorithms extract information about obstacles near pedestrians from top-down view images to improve the accuracy of trajectory prediction. However, mobile robots typically create local occupancy maps using LiDAR, rather than top-down view images. Meanwhile, the vision sensors on board robots provide egocentric view images, which contain fine-grained behavioral information about the pedestrians near the robot. To better use the information collected by LiDAR and on-board vision sensors, we propose MamMA, a Mamba-based pedestrian trajectory prediction algorithm considering occupancy maps and pedestrian awareness states. MamMA divides the occupancy map by patches and extracts obstacle features from each patch to create map features. Pedestrian awareness states are divided and considered, as some studies show that awareness states affect the perception and speed of pedestrians. Furthermore, a Mamba-based model is proposed to predict the future trajectories of pedestrians based on different types of features. Experiments on the STCrowd, SiT, JRDB, ETH, and UCY datasets show that MamMA achieves better average displacement error and final displacement error than the state-of-the-art algorithms. 

\end{abstract}

\section{INTRODUCTION}

Many pedestrian trajectory prediction algorithms have been proposed to help mobile robots avoid potential collisions during navigation. Although some algorithms achieve high prediction accuracy, the data required by these algorithms may not match the data collected by sensors equipped on the robots. 

\begin{figure}    
  \centering            
  \subfigure[The ETH dataset]   
  {
      \label{dataset:subfig1}\includegraphics[width=0.225\textwidth]{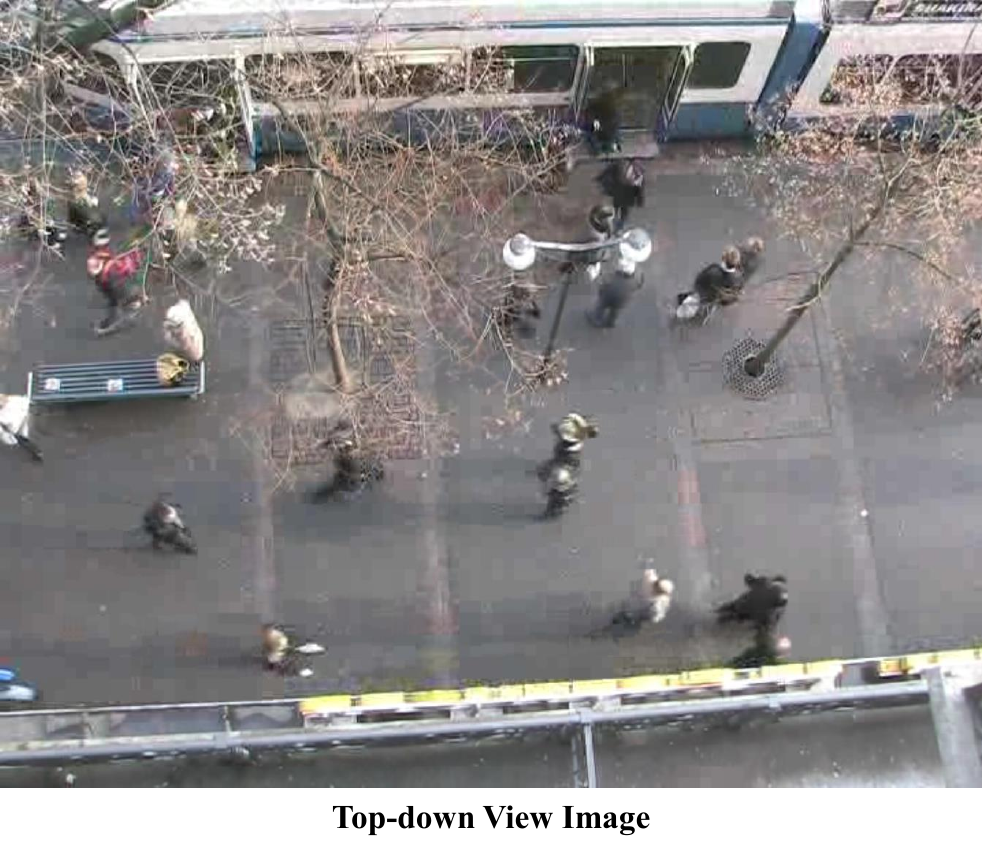}
  }
  \subfigure[The STCrowd dataset]
  {
      \label{dataset:subfig2}\includegraphics[width=0.225\textwidth]{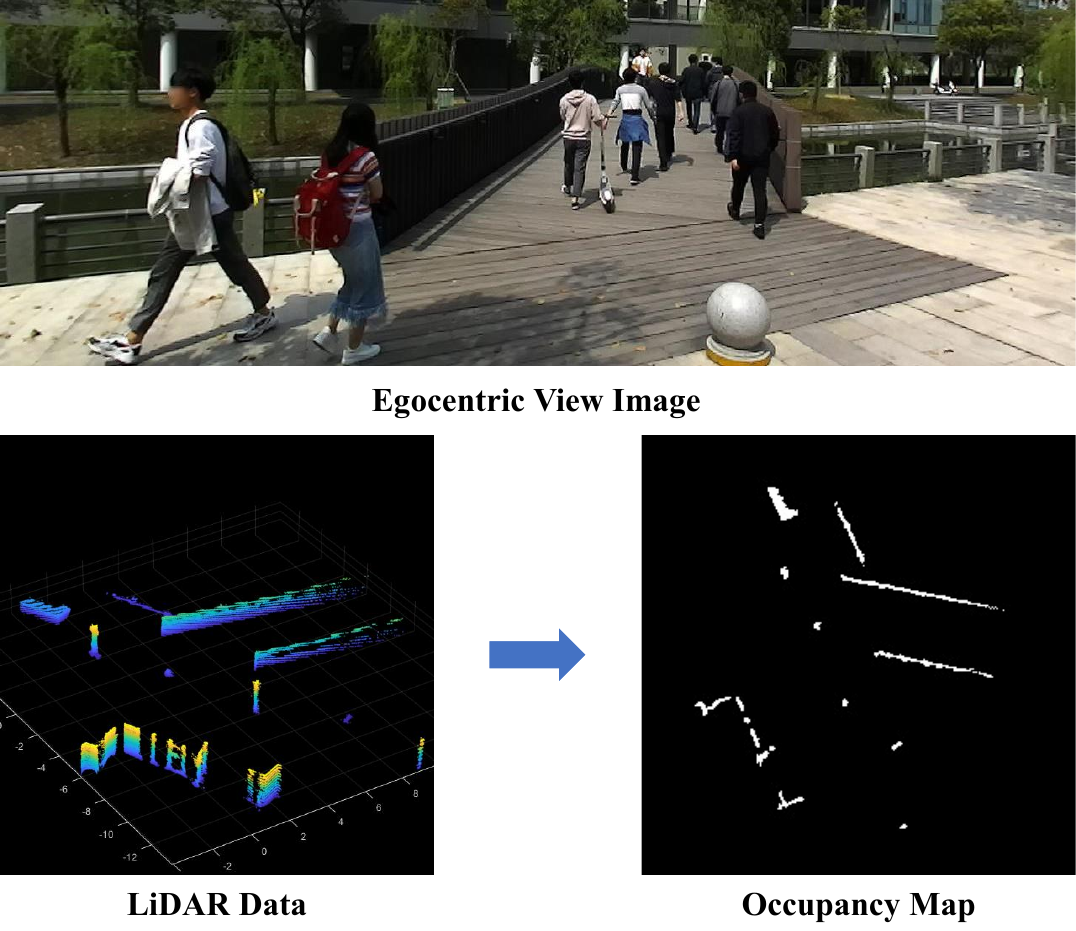}
  }
  \caption{Different types of pedestrian trajectory datasets.}
  \label{dataset}
\end{figure}

Most pedestrian trajectory prediction algorithms use datasets with top-down views, such as ETH~\cite{ETH} and UCY~\cite{UCY}, to train networks and evaluate performance. To obtain information about the environment for improving performance, some algorithms process top-down view images by semantic segmentation and deep neural networks to obtain features~\cite{SocialCVAE, ARGB2}. However, using top-down view images is not realistic for deploying robots in the real world. As shown in Fig.~\ref{dataset}, the mobile robot usually obtains egocentric view images with LiDAR data, rather than a top-down view image. Therefore, extracting environment features from LiDAR data can make the algorithm more suitable for real-world navigation.

\begin{figure}    
  \centering            
  \subfigure[Normal Pedestrian] 
  {
      \label{fig2:subfig1}\includegraphics[width=0.225\textwidth]{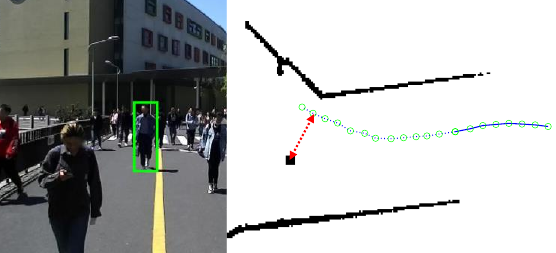}
  }
  \subfigure[Distracted Pedestrian] 
  {
      \label{fig2:subfig2}\includegraphics[width=0.225\textwidth]{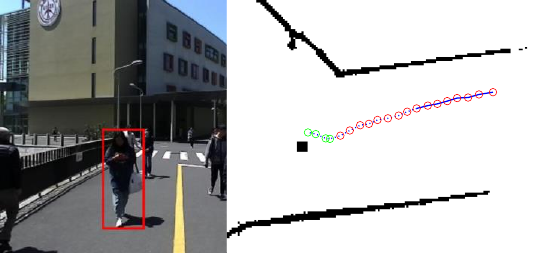}
  } 
  \caption{The trajectories of pedestrians with different awareness states in the STCrowd dataset. Green circles represent normal states, and red circles represent distracted states.}
  \label{fig2}
  
\end{figure}

In addition, as shown in Fig.~\ref{dataset}, the pedestrians in the egocentric view images are clearer than those in the top-down view images, which allows the algorithm to obtain more information about pedestrians, such as their awareness states. As shown in Fig.~\ref{fig2}, when using egocentric view images, we can divide pedestrian awareness states into normal and distracted, which can affect their trajectories. In Fig.~\ref{fig2}(a), the normal pedestrian always maintains a proper distance from obstacles by modifying its velocity. In contrast, the distracted pedestrian cannot detect obstacles until it is very close to them, then it has to modify velocity in a hurry, as shown in Fig.~\ref{fig2}(b).

To improve the algorithm's adaptability to real-world deployment and multi-dimensional understanding of pedestrians, we propose MamMA, a Mamba-based pedestrian trajectory prediction algorithm considering occupancy maps and pedestrian awareness states. MamMA uses the Mamba model~\cite{Mamba} to extract features from occupancy maps created by LiDAR data and considers the influence of pedestrians' awareness states on their trajectories. By using occupancy maps and pedestrian awareness states, MamMA has a better average displacement error (ADE) and final displacement error (FDE) than the state-of-the-art (SOTA) algorithms on the public dataset STCrowd~\cite{STCrowd}, SiT~\cite{SiT}, JRDB~\cite{JRDB}, ETH~\cite{ETH} and UCY~\cite{UCY}. 

The main contributions of this paper are as follows.
\begin{enumerate}

\item	We propose a Mamba-based scanning method that efficiently leverages occupancy maps in pedestrian trajectory prediction for extracting environment features.

\item   We propose an embedding method that effectively integrates the information of pedestrian awareness state and position for pedestrian trajectory prediction.

\item	We propose a Mamba-based pedestrian trajectory prediction network that extracts interaction features and trajectory features of pedestrians in temporal and spatial dimensions, as well as considers the effect of obstacles.

\end{enumerate}

\section{RELATED WORKS}
\subsection{Spatio-temporal Graph}
The spatio-temporal graph is used by many pedestrian trajectory prediction algorithms to describe pedestrian historical trajectories~\cite{STIGCN, IJCNN1, DSTIGCN}, because it can contain both temporal and spatial information about pedestrian trajectories. In the spatio-temporal graph, each node represents a pedestrian, and the correlations between different pedestrians are shown as edges~\cite{STGNE}. The connection between two pedestrians can be represented in many forms, such as the difference of their velocities in Social-STGCNN~\cite{ST-CNN1}. After the spatio-temporal graph is constructed, the temporal features and spatial features of the trajectories are extracted from the graph based on different methods, such as convolutional neural networks~\cite{SGCN}, recurrent neural networks~\cite{ST-RNN1}, and Transformer models~\cite{MRGT}. 

Our proposed method uses spatio-temporal graphs to describe the historical trajectories and interactions of pedestrians, and extracts features from the graphs based on Mamba models.

\subsection{Pedestrian Awareness and Trajectory}
Some studies have shown that the perception of pedestrians can be affected in some cases. ~\cite{phone0} indicates that the use of mobile phones reduces the perception abilities of pedestrians and reduces their walking speed. \cite{phone1} shows that pedestrians have increased reaction time and narrower perceptions of surroundings when using mobile phones. \cite{phone2} shows that pedestrians move significantly slower when using mobile phones. Such increased reaction time, narrowed perceptions and slower speed make significant differences in future trajectories of people with different awareness states, even though they have similar historical trajectories. 

\cite{phone3} shows that it is very common for pedestrians to use mobile phones, leading to low awareness states. However, almost all the existing pedestrian trajectory prediction algorithms do not consider the effect of pedestrian awareness states on their future trajectories. Our proposed method improves prediction accuracy by considering this effect.

\subsection{Dataset}
ETH~\cite{ETH} and UCY~\cite{UCY} are widely used in existing pedestrian trajectory prediction algorithms~\cite{SoicalI}. These datasets annotate the positions of pedestrians from top-down view videos and lack fine-grained information about pedestrians and LiDAR data. In recent years, some multi-modal datasets have been proposed, such as STCrowd~\cite{STCrowd}, SiT~\cite{SiT}, and JRDB~\cite{JRDB}, which provide both LiDAR data and egocentric view images, together with 2D and 3D position annotations of pedestrians. In the STCrowd dataset, the sensors are placed on a static bracket, and in the SiT and JRDB datasets, the sensors are carried by a mobile robot. 

Some algorithms use convolutional neural networks to extract information from top-down view images \cite{TITS3}. However, few existing algorithms use the information of pedestrian awareness states and occupancy maps, while our proposed approach is able to collect and use this information properly to improve prediction accuracy.

\begin{figure*}
  \begin{center}
  \includegraphics[width=7.1in]{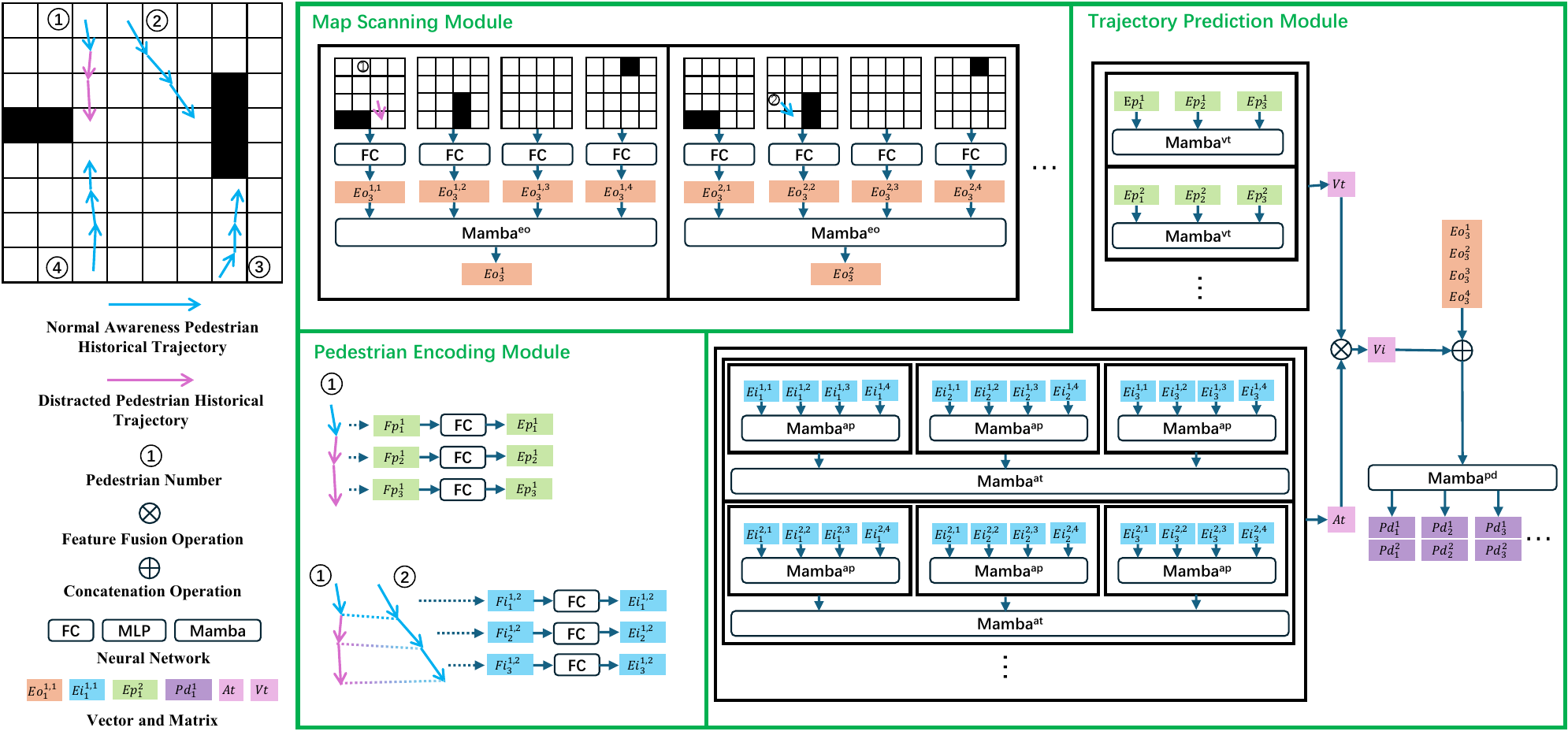}\\
  \caption{MamMA consists of three modules. The map scanning module extracts static obstacle features from the occupancy map. The pedestrian encoding module encodes the awareness state and positions of pedestrians. The trajectory prediction module fuses the features extracted by the other two modules and outputs the trajectory predictions. }
  \label{structure}
  \end{center}
\end{figure*}

\section{METHODOLOGY}

\subsection{Problem Formulation}

The information of the pedestrian $i$ at time $t$ is represented as $S_t^i = (P_t^i,D_t^i)$, where $P_t^i = (px_t^i,py_t^i)$ is the position and $D_t^i$ is the awareness state. The historical information $H_{t_{0}}^{i}$ and ground-truth future position $F_{t_{0}}^{i}$ of pedestrian $i$ at time $t_{0}$ can be represented as follows:

\begin{equation} 
H_{t_{0}}^{i} = {\left \{{ {S_{t_{0}}^{i}, S_{t_{0}-1}^{i},\ldots,S_{t_{0}-T_{obs}+1}^{i}} }\right\}}
\label{PF_obs1}
\end{equation}

\begin{equation} 
F_{t_{0}}^{i}  = {\left \{{ {P_{t_{0} +1}^{i}, P_{t_{0}+2}^{i}, \ldots,P_{t_{0} +T_{pred}}^{i}} }\right\}}
\label{PF_gt1}
\end{equation}

\noindent
where $T_{obs}$ and $T_{pred}$ are the time step numbers of the historical and predicted trajectories, respectively. Suppose there are $m$ pedestrians in the scene, then their historical and ground-truth future positions are denoted as $H_{t_0}^{1:m} = \{H_{t_0}^1, ..., H_{t_0}^m\}$ and $F_{t_0}^{1:m} = \{F_{t_0}^1, ..., F_{t_0}^m\}$, respectively. For the occupancy map, the positions of occupied cells are represented as a set of points $OBS={\left \{{ {O^{1} ,O^{2} ,\ldots, O^{n} } }\right\}}$, where $O^{i} = (ox^i,oy^i)$ is the position of the occupied cell $i$. MamMA inputs the $H_{t_{0}}^{1:m}$ and $OBS$, and outputs future trajectory predictions $\hat {F}_{t_{0}}^{1:m}$. 

\subsection{Algorithm Structure}
The model structure of MamMA is shown in Fig.~\ref{structure}, which consists of three modules: the pedestrian encoding module, the map scanning module, and the trajectory prediction module. 

The pedestrian encoding module is used to properly embed the information of pedestrian position and pedestrian awareness state. The map scanning module is used to extract information about obstacles in each patch of the occupancy map. The trajectory prediction module is used to predict the future trajectories of pedestrians based on the features of pedestrian historical trajectories, pedestrian interactions, and occupancy maps, from both temporal and spatial dimensions.

\subsection{Pedestrian Encoding Module}

In this module, the features of single pedestrian and two-pedestrian interaction are first calculated. For each pedestrian, a fully connected layer is used to embed its feature as follows:

\begin{equation} 
Fp^i_t = [P_t^i-P_{t_{0}-T_{obs}+1}^i,P_t^i-P_{t-1}^i, D_t^i], \quad Ep_t^i = \phi^{ep}(Fp_t^i) 
\label{Fp}
\end{equation}

\noindent
where $Fp^i_t$ is the information of the pedestrian $i$ at time $t$, including its displacement and awareness state, $\phi^{ep}$ is a fully connected layer with the PReLU activation function, whose dimensions of layers are $[6,n_{en}]$.

For the interaction of two pedestrians, we first determine their interaction type $I_t^{ij}$ based on their awareness states. The pedestrian interaction types include normal state to normal state, normal state to distracted state, and distracted state to all states, and each interaction type has a unique $I_t^{ij}$ code.

Then, a fully connected layer is used to embed the interaction features for each pedestrian pair as follows:

\begin{equation} 
Fi^{ij}_t = [P_t^i-P_t^j, \Delta P_t^i - \Delta P_t^j, I_t^{ij}], \quad Ei_t^{ij} = \phi^{ei}(Fi^{ij}_t) 
\label{Fp}
\end{equation}

\noindent
where $Fi^{ij}_t$ is the interaction feature between pedestrian $i$ and pedestrian $j$ at time $t$, including the difference in position and velocity and their interaction type, $\Delta P_t^i = P_t^i-P_{t-1}^i$, $\phi^{ei}$ is a fully connected layer with the PReLU activation function, whose dimensions of layers are $[7,n_{en}]$.

\subsection{Map Scanning Module}

With the map scanning module, the influence of static obstacles on pedestrian trajectories can be considered. Suppose the map is divided into $n_p$ patches with an edge length of $d_p$, and up to $n_o$ obstacles in each patch. For patch $k$, the position set of obstacles in it can be represented as $OBS^k={\left \{{ {O^{k_1} ,O^{k_2} ,\ldots, O^{k_{n_o}} } }\right\}}$. 

We first generate $Eo_t^{i,k_j}$, the influence features of obstacle $j$ in patch $k$ to pedestrians $i$ at time $t$. Two fully connected layers are used to embed the relative positions and velocities of obstacles to the pedestrian, and $Eo_t^{i,k_j}$ is calculated by the Hadamard product. The process is shown as follows:

\begin{equation} 
Eop_t^{i,k_j} = \phi^{eop}(P_t^i-O^{k_j}), \quad  Eov_t^{i,k_j} = \phi^{eov}(\Delta P_t^i) 
\label{Eop}
\end{equation}

\begin{equation} 
Eo_t^{i,k_j} = {Eop_t^{i,k_j}} \odot {Eov_t^{i,k_j}} 
\label{Eo}
\end{equation}

\noindent
where $\phi^{eop}$ and $\phi^{eov}$ are the fully connected layers with the PReLU activation functions, whose dimensions of layers are both $[2,n_{eo}]$, $\odot$ represents the Hadamard product.

Then, we use a fully connected layer to combine the features of all obstacles in patch $k$ to generate the patch feature $Eo_t^{i,k}$. The calculation of $Eo_t^{i,k}$ is shown as follows:

\begin{equation} 
{Eo_t^{i,k}} = \phi^{eok}([{Eo_t^{i,k_1}},{Eo_t^{i,k_2}},\ldots,{Eo_t^{i,{k_{n_o}}}}]) 
\label{Eo}
\end{equation}

\noindent
where $\phi^{eok}$ is the fully connected layer with the PReLU activation function, whose dimensions of layers are $[n_{eo}*n_{o},n_{en}]$.

Finally, we use a Mamba model to scan all patches of the map and generate $Eo_t^{i}$, the map features for pedestrians $i$ at time $t$. The process is shown as follows:

\begin{equation} 
Eo_t^{i} = Mam^{eo}(Eo_t^{i,(1:n_p)}) 
\label{Vt}
\end{equation}

\noindent
where $Mam^{eo}$ is a Mamba model with state dimension of $n_s$, and $Eo_t^{i,(1:n_p)}$ represents a feature sequence from $Eo_t^{i,1}$ to $Eo_t^{i,n_p}$. The map scanning module generates $Eo_{t_{0}}^i$ for all patches with all pedestrians at time $t_{0}$.

\subsection{Trajectory Prediction Module}

The trajectory prediction module is used to combine the information obtained from the former two modules and output the trajectory prediction. A spatio-temporal graph is first created, which has a node matrix $V$ with dimensions of $[T_{obs},m,n_{en}]$ and an adjacency matrix $A$ with dimensions of $[T_{obs},m, m,n_{en}]$. The assignment rules of $V$ and $A$ are $V_{t,i,:}= Ep_t^{i}$ and $A_{t,i,j,:}= Ei_t^{ij}$.

Then, Mamba models are used to extract features from $V$ and $A$ along both temporal and spatial dimensions. For $V$, features are extracted along the temporal dimension. For $A$, features are extracted using Mamba models along spatial and then temporal dimensions. The process is shown as follows: 

\begin{equation} 
Vt_{t,i,:} = Mam^{vt}(V_{1:t,i,:}) 
\label{Vt}
\end{equation}

\begin{equation} 
Ap_{t,i,j,:} = Mam^{ap}(A_{t,i,1:j,:}) 
\label{Ap}
\end{equation}

\begin{equation} 
At_{t,i,j,:} = Mam^{at}(Ap_{1:t,i,j,:}) 
\label{At}
\end{equation}

\noindent
where $Mam^{vt}$, $Mam^{ap}$, and $Mam^{at}$ are Mamba models with state dimension of $n_s$. $Vt$ and $At$ are matrices after feature extraction and have the same dimensions as $V$ and $A$, respectively.

Then $Vi$ is calculated by fusing the features contained in $Vt$ and $At$. $Vi$ has the same dimensions as $V$ and considers the influence of other pedestrians and static obstacles. The calculation rule of $Vi$ is as follows: 

\begin{equation} 
 Vi_{t,i,k} = Vt_{t,i,k} + {\sum _{j=1}^{m} \sum _{c=1}^{n_{en}} Vt_{t,j,k} \cdot At_{t,i,j,c}} 
\label{Vi}
\end{equation}

After the batch normalization of $Vi$, we concatenate $Vi$ and $Eo_{t_{0}}^i$ as $Vo$, and use a multilayer perceptron (MLP) to process it along the first dimension to obtain the matrix $Vd$ for decoding. The process is shown as follows: 

\begin{equation} 
Vd = \psi^{vd}(Vo) 
\label{Vd}
\end{equation}

\noindent
where $\psi^{vd}$ is a two-layer MLP with PReLU activation functions, whose dimensions of layers are $[T_{obs}+1, T_{pred}, T_{pred}]$. $Vo$ has dimensions of $[T_{obs}+1,m,n_{en}]$, while $Vd$ has dimensions of $[T_{pred},m,n_{en}]$.

Finally, a Mamba model and an MLP are used to decode $Vo$ and output the predicted displacement of pedestrian $i$ at time $t$. The process is shown as follows: 

\begin{equation} 
Pd^i_t = Mam^{pd}(Vd_{1:t,i,:}), \quad  \Delta \hat P^i_t = \psi ^{p}(Pd^i_t) 
\label{pd}
\end{equation}

\begin{equation} 
\hat P^i_t = \hat P^i_{t-1} + \Delta \hat P^i_t \quad {(t=1,2,\ldots,T_{pred})}  
\label{RX}
\end{equation}

\noindent
where $Mam^{pd}$ is a Mamba model with state dimension of $n_s$, and $\psi^{p}$ is a two-layer MLP with PReLU activation functions, whose dimensions of layers are $[n_{en}, n_{en}, 2]$. $\Delta \hat P^i_t$ and $\hat P^i_t$ are the displacement and position predictions of pedestrian $i$ at time $t$, respectively.

\section{EXPERIMENTS AND ANALYSIS}

\subsection{Experiment Settings}

\begin{table*}
\centering
\setlength{\tabcolsep}{4.5pt}
\caption{The results for the STCrowd, SiT and JRDB datasets. \textbf{Bold} and \underline{underline} mark the best and second-best results, respectively.}
\label{STC_result}

\begin{tabular}{ccccccccccccc}
\toprule 
Dataset & \multicolumn{4}{c}{STCrowd} & \multicolumn{4}{c}{SiT} & \multicolumn{4}{c}{JRDB} \\
\cmidrule(l){2-5} \cmidrule(l){6-9} \cmidrule(l){10-13} 
Metric & \multicolumn{3}{c}{$\text{minADE}_{3}$/$\text{minFDE}_{3}$  $\downarrow$} & {$p_{coll}$  $\downarrow$} & \multicolumn{3}{c}{$\text{minADE}_{3}$/$\text{minFDE}_{3}$  $\downarrow$} & {$p_{coll}$  $\downarrow$} & \multicolumn{3}{c}{$\text{minADE}_{3}$/$\text{minFDE}_{3}$  $\downarrow$} & {$p_{coll}$  $\downarrow$} \\
 \cmidrule(l){2-4} \cmidrule(l){5-5} \cmidrule(l){6-8} \cmidrule(l){9-9} \cmidrule(l){10-12} \cmidrule(l){13-13} 
 Sample & \text{All}  & $S_{dis}$  & $S_{obs}$  &  \text{All} & \text{All}  & $S_{dis}$  & $S_{obs}$  &  \text{All} & \text{All}  & $S_{dis}$  & $S_{obs}$  &  \text{All} \\
\midrule
            Social-STGCNN    & 0.68/1.18 & 0.63/1.09 & 0.80/1.43 & 4.14\% & 0.62/1.09 & 0.85/1.47 & 0.64/1.14 & 8.86\% & 0.87/1.50  & 1.28/2.40 & 0.98/1.71 & 2.49\% \\
            SGCN    & \underline{0.38}/\underline{0.69} & 0.43/\underline{0.81} & 0.49/\underline{0.92} & 2.49\% & 0.44/0.81 & 0.74/1.35 & 0.47/0.87 & 5.09\% & \underline{0.68}/\underline{1.23} & \underline{0.94}/1.91 & 0.89/1.52 & 1.92\% \\
            Social-Implicit     & \underline{0.38}/0.76 & \underline{0.42}/0.86 & \underline{0.48}/0.99 & \underline{1.72\%}  & 0.44/0.85 & \underline{0.64}/1.20 & 0.47/0.92 & 4.66\% & 0.69/1.35 & 1.14/2.35 & 0.83/1.64 & \underline{1.56\%} \\ 
            MRGT             & 0.39/0.77 & 0.43/0.84 & 0.49/0.99 & 2.23\% & 0.44/0.86 & \underline{0.64}/1.22 & 0.47/0.92 & 6.63\% & \underline{0.68}/1.32 & 1.03/2.12 & \underline{0.80}/1.54 & 1.63\% \\
            IMGCN             & 0.42/0.76 & 0.48/0.86 & 0.55/1.02 & 2.30\% & 0.45/0.81 & 0.69/1.25 & 0.47/0.87 & 4.68\% & 0.71/1.28 & 0.95/\underline{1.90} & 0.89/1.56 & 2.04\% \\
            STIGCN             & 0.39/0.71 & 0.45/0.86 & 0.51/0.96 & 2.25\% & 0.44/0.82 & \underline{0.64}/\underline{1.16} & 0.46/0.86 & 4.51\% & 0.69/1.24 & 0.95/1.92 & 0.91/1.55 & 1.85\% \\
            DSTIGCN          & 0.41/0.72 & 0.47/0.82 & 0.53/0.97 & 2.41\% & \underline{0.42}/\underline{0.77} & 0.67/1.23 & \underline{0.44}/\underline{0.83} & \underline{4.22\%} & 0.71/1.26 & 0.96/1.96 & 0.83/\underline{1.48} & 1.93\% \\
            MamMA    &  \textbf{0.27}/\textbf{0.49} & \textbf{0.29}/\textbf{0.51} & \textbf{0.32}/\textbf{0.59} & \textbf{0.87\%} &  \textbf{0.36}/\textbf{0.68} & \textbf{0.55}/\textbf{1.05} & \textbf{0.38}/\textbf{0.74} & \textbf{2.63\%} &  \textbf{0.55}/\textbf{1.05} & \textbf{0.76}/\textbf{1.59} & \textbf{0.66}/\textbf{1.27} & \textbf{1.03\%} \\
\bottomrule
\end{tabular}
\end{table*}

We compare MamMA with the following SOTA algorithms: DSTIGCN\cite{DSTIGCN} (2025), STIGCN\cite{STIGCN} (2024), IMGCN\cite{IMGCN} (2024), MRGT\cite{MRGT} (2023), Social-Implicit\cite{SoicalI} (2022), SGCN\cite{SGCN} (2021), Social-STGCNN\cite{ST-CNN1} (2020).  For the STCrowd~\cite{STCrowd}, SiT~\cite{SiT} and JRDB~\cite{JRDB} datasets, as test set labels are not provided, we train all the algorithms on the training set, and evaluate them on the validation set. For the ETH~\cite{ETH} and UCY~\cite{UCY} datasets, we train MamMA on the training set and report the test set results for the weights with the best validation set performance, and we use the weights provided by the authors for the SOTA algorithms.

To evaluate the performance of the algorithm in more detail, we further divide the samples into All, $S_{dis}$, and $S_{obs}$ for the STCrowd, SiT, and JRDB datasets. The samples of pedestrians who show distracted states during the observation time are denoted as $S_{dis}$. The samples whose ground-truth future trajectories $F_{t}^{i}$ are less than 1 meter away from the nearest obstacle are denoted as $S_{obs}$, which have potential interaction with static obstacles. We annotate the pedestrian awareness states for the STCrowd, SiT, and JRDB datasets.

The loss function of MamMA minimizes the average displacement error (ADE) between the prediction $\hat {F}_{t_{0}}^{1:m}$ and the ground-truth trajectory $F_{t_{0}}^{1:m} $. ADE and FDE are used as metrics in experiments, whose calculation methods are shown in~\cite{ST-CNN1}. Referring to~\cite{bestk3}, for MamMA and the SOTA algorithms, we output 3 trajectories and select the trajectory with the best ADE and FDE to calculate metrics, that is, $\text{minADE}_{3}$ and $\text{minFDE}_{3}$. In addition, we calculate the proportion of trajectory predictions that collide with static obstacles among all predictions, denoted as $p_{coll}$. $p_{coll}$ shows the rationality of the algorithm's predictions, as the ground-truth trajectories do not collide with static obstacles.

We use LiDAR data to generate occupancy grid maps with a resolution of 0.1 meters. The length and width of the map are 48 meters and 30 meters, respectively. We set $d_p = 3$ when dividing the patches, so $n_p = 160$. The structural parameters of the network are set as $n_{eo}=16$, $n_{o}=75$, $n_{en}=64$, $n_s=64$. The learning rate is set to 0.001, the batch size is set to 8, and the epoch number is set to 120. We use the same observation time and prediction time settings as the SOTA algorithms, that is, 3.2 seconds for observation and 4.8 seconds for prediction. All the datasets annotate the data at a rate of 2.5 FPS, so $T_{obs}=8$ and $T_{pred}=12$.

\begin{table}
\centering
\caption{The model parameters, model size, and inference time of the algorithms for the STCrowd dataset.}
\label{predtime}

\begin{tabular}{cccc}
\toprule
& 
parameters  & 
size  & 
inference time \\
\midrule
Social-STGCNN    & 7563 & 44 KB & 1.34 ms \\
SGCN		     & 25369 & 144 KB  & 3.04 ms \\
Social-Implicit	 & 5836 & 72 KB  & 1.22 ms\\
MRGT	         & 4358926 & 17061 KB & 4.45  ms\\
IMGCN	         & 23384 & 104 KB & 3.37  ms\\
STIGCN	         & 27564 & 125 KB & 2.12  ms \\
DSTIGCN	         & 17767 & 97 KB & 7.41 ms \\
MamMA         & 347333 & 1390 KB & 2.64 ms\\

\bottomrule
\end{tabular}
\end{table}

\subsection{Quantitative Experiments and Analysis}

As shown in Table~\ref{STC_result}, MamMA has the best $\text{minADE}_{3}$, $\text{minFDE}_{3}$, and $p_{coll}$ for the STCrowd, SiT and JRDB datasets, which shows that MamMA is well adapted to data collected by real-world robots.

For the STCrowd dataset, the $\text{minADE}_{3}$ and $\text{minFDE}_{3}$ for $S_{dis}$ of the SOTA algorithms except for Social-STGCNN rise by over 0.04 meters and 0.1 meters compared to those for all samples, while those of MamMA rise by only 0.02 meters. The $\text{minADE}_{3}$ and $\text{minFDE}_{3}$ for $S_{obs}$ of the SOTA algorithms rise by over 0.1 meters and 0.18 meters compared to those for all samples, respectively, while those of MamMA rise by only 0.05 meters and 0.1 meters. For the SiT and JRDB datasets, the performance degradation of MamMA for $S_{dis}$ and $S_{obs}$ is less than that of the almost SOTA algorithms. The results show that MamMA better considers the effects of obstacles and awareness states on pedestrians than the SOTA algorithms.

In addition, the $p_{coll}$ of MamMA is significantly lower than that of the SOTA algorithms. The $p_{coll}$ of MamMA is almost less than half of the SOTA algorithms in the STCrowd dataset, and less than two-thirds of the SOTA algorithms in the SiT and JRDB datasets. MamMA's lower $p_{coll}$ means the trajectories predicted by MamMA are less likely to hit obstacles. 

\begin{figure}    
  \begin{center}
  \includegraphics[width=0.48\textwidth]{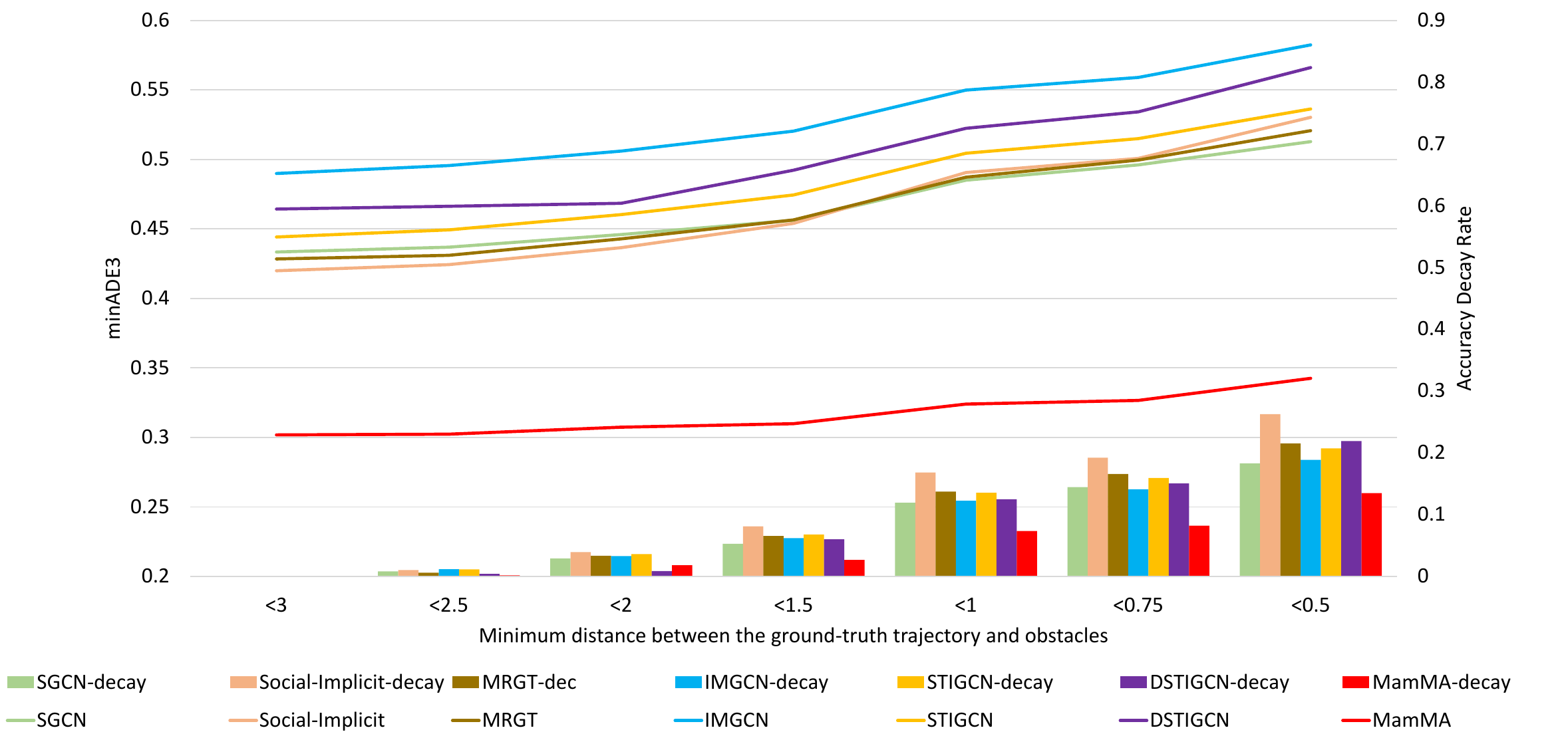}\\
  \caption{The $\text{minADE}_{3}$ and prediction accuracy decay ratio of MamMA and some SOTA algorithms when the minimum distance between the ground-truth trajectory and obstacles satisfies different conditions. “$<3$” denotes the set of all the samples where the minimum distance between the ground-truth trajectory and obstacles is less than 3 meters.}
  \label{fig6}
  \end{center}
\end{figure}

When the ground-truth future trajectory of a pedestrian is close to obstacles, accurately predicting this trajectory becomes difficult because it is more likely to be affected by obstacles. To study the effect of obstacles on the prediction accuracy of the algorithms, we calculated the $\text{minADE}_{3}$ of the algorithms when the minimum distance between the ground-truth trajectory and obstacles satisfies different conditions. The results are shown in Fig.~\ref{fig6}. 

When the minimum distance between the ground-truth trajectory and obstacles decreases from 3 meters to 0.5 meters, the $\text{minADE}_{3}$ of MamMA increases by only about 0.04 meters, and prediction accuracy decreases by only about 13.5\%, which is the smallest decay amount and decay ratio among the algorithms. For SOTA algorithms, even the SGCN, the least affected algorithm, has a $\text{minADE}_{3}$ increase of about 0.08 meters and an accuracy decay ratio of about 18.3\%. The results indicate that the prediction accuracy of MamMA is least affected by obstacles.

Table~\ref{predtime} shows the model parameters, the model size, the average inference time of the algorithms. The inference time is calculated based on RTX 5090 and Ryzen 7 7700. Although MamMA has a relatively large model size, it still has an acceptable inference time compared to the SOTA algorithms.

\subsection{Qualitative Analysis}

\begin{figure*}
  \begin{center}
  \includegraphics[width=6.6in]{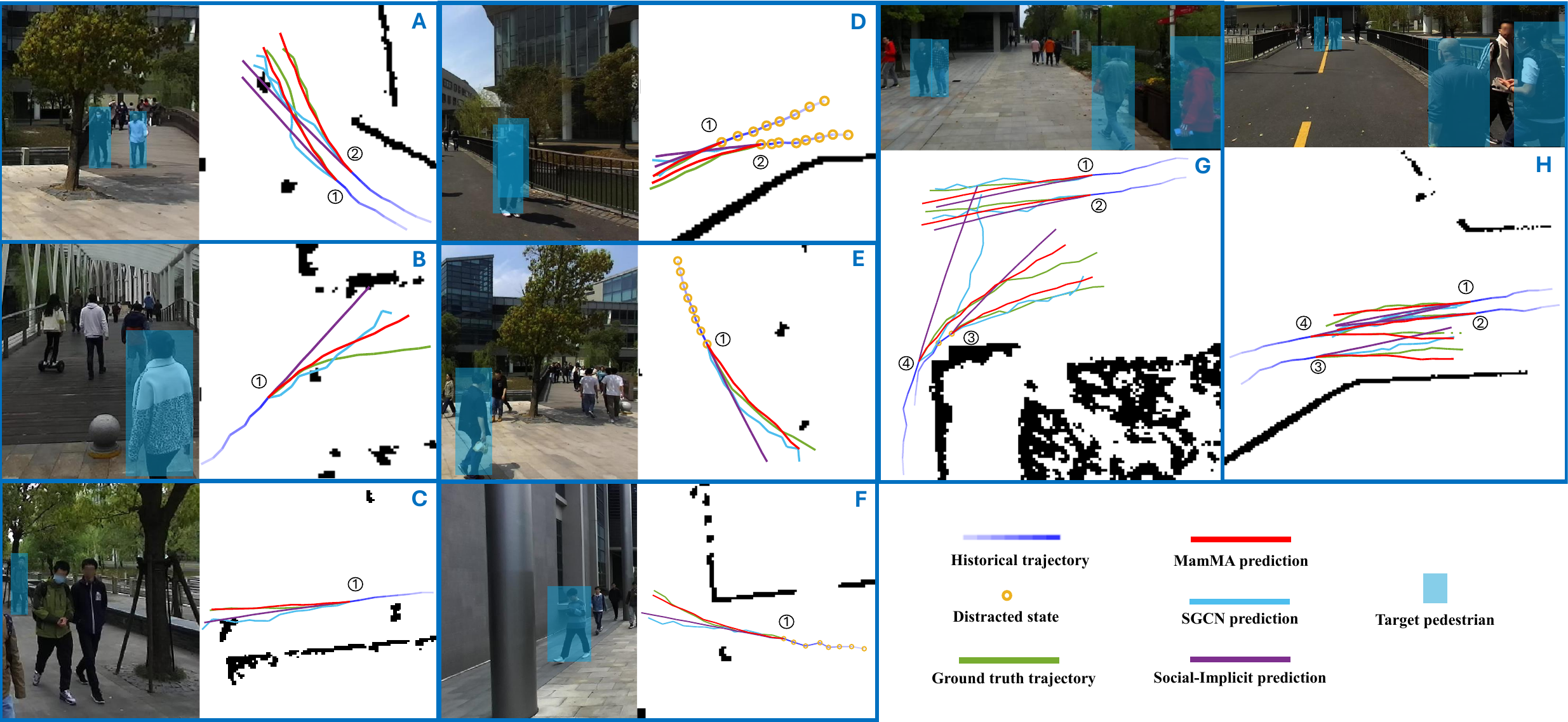}\\
  \caption{The prediction results of MamMA and some SOTA algorithms. }
  \label{visi}
  \end{center}
\end{figure*}

The prediction results of some scenes are shown in Fig.~\ref{visi}. MamMA makes trajectory predictions more accurate and less likely to hit static obstacles by scanning maps with Mamba models. In scenes A, B, and C, some trajectories predicted by the SOTA algorithms collide with static obstacles.

MamMA also accurately predicts the future trajectory of pedestrians who interact with other pedestrians. In scene G, when the four pedestrians meet at the intersection, MamMA correctly predicts that pedestrian 1 and pedestrian 2 go straight but pedestrian 3 and pedestrian 4 turn right, while the trajectories predicted by SGCN and Social-Implicit collide. Also, in scene H, when the four pedestrians meet head-on, MamMA correctly predicts that pedestrian 3 and pedestrian 4 turn to avoid pedestrian 1 and pedestrian 2, while SGCN and Social-Implicit fail to do that.

MamMA achieves accurate trajectory prediction of distracted pedestrians by encoding the pedestrian awareness. MamMA accurately predicts that the future trajectories of the two distracted pedestrians are closer to each other in scene D, and properly predicts that the future trajectories of distracted pedestrians are closer to static obstacles in scenes E and F.

The qualitative analysis indicates that MamMA has a higher prediction accuracy than the SOTA algorithm and can effectively take into account the effects of static obstacles and awareness states on the future trajectories of pedestrians.  

\begin{table}
\centering
\caption{The ablation study results on STCrowd. \textbf{Bold} and \underline{underline} mark the best and second-best results, respectively.}
\label{result3}
\begin{tabular}{ccccc}
\toprule 
Metric & \multicolumn{3}{c}{$\text{minADE}_{3}$/$\text{minFDE}_{3}$ \ $\downarrow$} & {$p_{coll}$ \ $\downarrow$}\\
 \cmidrule(l){2-4} \cmidrule(l){5-5} 
 Sample & \text{All}  & $S_{dis}$  & $S_{obs}$  &  \text{All} \\
\midrule
            Mam   & 0.32/0.61 & 0.38/0.73  & 0.39/0.74 & 1.83\% \\
	        MamM    & \underline{0.28}/\underline{0.50} & \underline{0.31}/\underline{0.54} & \underline{0.34}/\underline{0.62} & \underline{1.11\%} \\
            MamMA    &  \textbf{0.27}/\textbf{0.49} & \textbf{0.29}/\textbf{0.51} & \textbf{0.32}/\textbf{0.59} & \textbf{0.87\%} \\
\bottomrule
\end{tabular}
\end{table}

\begin{table}
\centering
\caption{The ablation study results on STCrowd. \textbf{Bold} and \underline{underline} mark the best and second-best results, respectively.}
\label{result4}
\begin{tabular}{ccc}
\toprule 
Metric & {$\text{minADE}_{3}$/$\text{minFDE}_{3}$ \ $\downarrow$} & {$p_{coll}$ \ $\downarrow$}\\
 \cmidrule(l){2-2} \cmidrule(l){3-3} 
 Sample & \text{All}  &  \text{All} \\
\midrule
            MamMA w/o $Mam^{vt}$   & \underline{0.29}/\underline{0.54} & \underline{1.15\%}  \\
	        MamMA w/o $Mam^{ap}$   & 0.30/0.56 & 1.19\% \\
            MamMA w/o $Mam^{at}$   & 0.31/0.56 & 1.18\% \\
            MamMA w/o $Mam^{pd}$   & 0.30/\underline{0.54} & 1.25\%  \\
            MamMA w/o $Mam^{vt}$, $Mam^{pd}$   & 0.32/0.59 & 1.40\%  \\
            MamMA w/o $Mam^{ap}$, $Mam^{at}$   & 0.51/0.89 & 1.92\%  \\
            MamMA w/o Mamba   & 0.88/1.60 & 3.27\%  \\
            MamMA    &  \textbf{0.27}/\textbf{0.49} & \textbf{0.87\%} \\
\bottomrule
\end{tabular}
\end{table}

\subsection{Ablation Study}

\begin{table}
\centering
\caption{The ablation study results on the ETH and UCY datasets. \textbf{Bold} and \underline{underline} mark the best and second-best results, respectively.}
\label{result5}
\begin{tabular}{ccccc}
\toprule 
Metric & \multicolumn{4}{c}{$\text{minADE}_{3}$/$\text{minFDE}_{3}$ \ $\downarrow$} \\
 \cmidrule(l){2-5} 
Dataset & HOTEL  &
UNIV &
ZARA1  &
ZARA2  \\
\midrule
Social-STGCNN	 & 0.60/1.10 & 0.62/1.18 & 0.48/0.87 & 0.43/0.77  \\
SGCN		     & 0.45/0.85 & 0.49/0.98 & 0.40/0.78 & \underline{0.29}/\underline{0.59}  \\
Social-Implicit  & 0.47/0.89  & 0.51/1.06 & 0.41/0.83 & 0.37/0.70 \\   
MRGT             & \underline{0.41}/\underline{0.77} & 0.64/1.25 & 0.42/0.84 & 0.39/0.78\\
IMGCN            & 0.46/0.83 & 0.48/0.97 & 0.41/0.81 & 0.34/0.66\\
STIGCN           & 0.43/0.83 & 0.52/1.00 & 0.39/0.76 & 0.50/0.97\\
DSTIGCN          & 0.46/0.91 & \underline{0.46}/\underline{0.92} & \underline{0.36}/\underline{0.72} & 0.30/0.60\\
Mam  & \textbf{0.31}/\textbf{0.59} & \textbf{0.45}/\textbf{0.91}  & \textbf{0.32}/\textbf{0.66} & \textbf{0.25}/\textbf{0.54}\\
\bottomrule
\end{tabular}
\end{table}

We first study the effect of using maps and pedestrian awareness states on the performance of MamMA by Table~\ref{result3}, where MamM denotes that MamMA only uses occupancy maps, and Mam denotes MamMA without using occupancy maps and pedestrian awareness states. 

As shown in Table~\ref{result3}, MamMA has better performance than the other algorithms due to the use of maps and awareness state information. The $\text{minADE}_{3}$, $\text{minFDE}_{3}$ and $p_{coll}$ of Mam are worse than those of MamM, due to the lack of using maps. Due to the lack of using awareness states, $\text{minADE}_{3}$ and $\text{minFDE}_{3}$ for $S_{dis}$ of MamM are worse than those of MamMA. The results show that using maps and pedestrian awareness states can effectively improve the performance of MamMA.

Then, we study the effect of different models in MamMA by Table~\ref{result4}, where MamMA w/o $Mam^{vt}$ denotes MamMA without using $Mam^{vt}$, namely let $Vt_{t,i,:} = V_{t,i,:}$, and MamMA w/o Mamba denotes MamMA without using $Mam^{vt}$, $Mam^{ap}$, $Mam^{at}$ and $Mam^{pd}$.

As shown in Table~\ref{result4}, when any one of the Mamba models is removed, the $\text{minADE}_{3}$ and $\text{minFDE}_{3}$ of MamMA increase by at least 0.02 and 0.05 meters, respectively. The absence of both $Mam^{vt}$ and $Mam^{pd}$ causes some performance degradation of MamMA, and the absence of both $Mam^{ap}$ and $Mam^{at}$ makes the performance of MamMA worse than almost all the SOTA algorithms. The results indicate the positive impact of Mamba models on the network performance. 

Further, we compare the performance of algorithms on the ETH and UCY datasets. These datasets do not provide LiDAR data, and pedestrian awareness states are difficult to annotate because pedestrians are not clear enough in the top-down view videos. Therefore, we use Mam rather than MamMA, as MamMA has to use occupancy maps and pedestrian awareness states. Table~\ref{result5} shows the prediction performance of the algorithms on the ETH and UCY datasets. Despite the lack of occupancy maps and awareness states, Mam still achieves the best performance in the ETH and UCY datasets, which shows that our method still has good performance even without map and awareness state information. This result is consistent with the results shown in Tables~\ref{STC_result} and~\ref{result3}, where Mam performs better than the SOTA algorithms for the STCrowd dataset.

\section{CONCLUSION}

In this paper, we propose MamMA, a Mamba-based trajectory prediction algorithm that leverages occupancy maps (derived from LiDAR) and pedestrian awareness states (extracted from egocentric images). By utilizing data collected by on-board sensors, MamMA achieves better adaptation to real-world robot deployments and improves prediction accuracy.

We propose an encoding method to combine pedestrian trajectory information and pedestrian awareness information, and propose a method that uses the Mamba model to extract information about obstacles from occupancy maps. MamMA uses a Mamba-based model to extract pedestrian trajectory features and interaction features in both temporal and spatial dimensions and predict future trajectories. Experiments on the STCrowd, SiT, JRDB, ETH and UCY datasets show that MamMA has better performance and similar inference time compared to the SOTA algorithms.

\bibliographystyle{IEEEtran}
\bibliography{Bibliography}

\end{document}